\documentclass[10pt,twocolumn,letterpaper]{article}

\usepackage{cvpr}
\usepackage{times}
\usepackage{epsfig}
\usepackage{graphicx}
\usepackage{amsmath}
\usepackage{amssymb}
\usepackage{enumitem}
\usepackage{multirow}

\usepackage[breaklinks=true,bookmarks=false]{hyperref}

\cvprfinalcopy % *** Uncomment this line for the final submission

\def\cvprPaperID{****} % *** Enter the CVPR Paper ID here

\begin{document}

%%%%%%%%% TITLE
%\title{Single Image Dehazing Using Perceptual Losses}

\title{PAD-Net: A Perception-Aided Single Image Dehazing Network}
\author{Yu Liu\textsuperscript{1} and Guanlong Zhao\textsuperscript{2}\\
\textsuperscript{1}Department of Electrical and Computer Engineering, Texas A\&M University\\
\textsuperscript{2}Department of Computer Science and Engineering, Texas A\&M University\\
{\tt\small \{yliu129, gzhao\}@tamu.edu}
}

\maketitle
%\thispagestyle{empty}

%%%%%%%%% ABSTRACT
\begin{abstract}
In this work, we investigate the possibility of replacing the $\ell_2$ loss with perceptually derived loss functions (SSIM, MS-SSIM, etc.) in training an end-to-end dehazing neural network. Objective experimental results suggest that by merely changing the loss function we can obtain significantly higher PSNR and SSIM scores on the SOTS set in the RESIDE dataset, compared with a state-of-the-art end-to-end dehazing neural network (AOD-Net) that uses the $\ell_2$ loss. The best PSNR we obtained was 23.50 (4.2\% relative improvement), and the best SSIM we obtained was 0.8747 (2.3\% relative improvement.)
\end{abstract}

%%%%%%%%% BODY TEXT
\section{Introduction}
\label{Intro}
Due to the existence of air pollution, dust, mist, and fumes, images taken in an outdoor environment will often contain complicated, non-linear and data-dependent noises, known as haze, which challenges many high-level computer vision tasks such as object detection and recognition. Taking autonomous-driving as an example, hazy or foggy weather will obscure the vision of on-board cameras and create a loss of contrast in the subject with light scattering through the haze particles, adding superior difficulties for self-driving tasks. Thus, dehazing is a highly desirable image restoration technique to enhance better results of computational photography and computer vision tasks.

Early approaches of dehazing often require additional information such as scene depth to be given or captured from comparing multiple different images of the same scene~\cite{tan2000enhancement,schechner2001instant,kopf2008deep}. While these methods can effectively enhance the visibility of hazy images, their tractability is limited since the required additional information or multiple images are not always available in practice. 

To address this problem, a single-image dehazing system, which aims at restoring the underlying clean image from a observed hazy image, is more feasible for real application and has received an increased interests in recent years. Traditional single-image dehazing methods exploit natural image prior and perform statical analysis~\cite{he2011single,tang2014investigating,zhu2015fast,berman2016non}. More recently, dehazing algorithms based on neural networks \cite{cai2016dehazenet,ren2016single,li2017aod} have shown state-of-the-art performance, among which the AOD-Net \cite{li2017aod} has the ability to train an end-to-end system while outperforming the others on multiple evaluation metrics. AOD-Net minimizes the $\ell_2$ norm of the difference between the haze and clean images. %However, the $\ell_2$ norm suffers from a few known issues that may harness the performance for a maximum optimized and perceptual clean image. First, $\ell_2$ norm assumes a white Gaussian noise, which is not suitable for the dehazing task. Second, $\ell_2$ norm does not correlate well with the human perception of image quality \cite{zhang2012comprehensive}. Last, $\ell_2$ treats the impact of noise on the local characteristics of an image independently. However, according to \cite{wang2004image}, the sensitivity of the Human Visual System to noise depends on the local structure of a vision. 
 However, the $\ell_2$ norm suffers from a handful of known limitations that may leave the dehazed image output of the AOD-Net away from the optimal quality, especially considering about its correlation with human perception of image quality\cite{zhang2012comprehensive}. On the one hand, $\ell_2$ norm implicitly assumes a white Gaussian noise, which is an oversimplified case that is not valid in general dehazing cases. On the other hand, $\ell_2$ treats the impact of noise independently to the local characteristics, such as structural information, luminance and contrast, of an image. However, according to \cite{wang2004image}, the sensitivity of the Human Visual System (HVS) to noise depends on the local properties and structure of a vision. 

Alternatively, the structural similarity index (SSIM), is widely employed as a metric to evaluate image processing algorithms from a more perceptual point of view. Besides, it also possesses a differentiable property and can be used as a cost function. Therefore, in this work, inspired by \cite{zhao2017loss}, we propose to use loss functions that match with human perception (e.g., SSIM \cite{wang2004image}, MS-SSIM \cite{wang2003multiscale}) as training objectives of a dehazing neural network developed based on the AOD-Net~\cite{li2017aod}. We call this \underline{P}erception-\underline{A}ided Single Image \underline{D}ehazing Network: \textit{PAD-Net}. We hypothesize that even without changing the neural network architecture, the PAD-Net will lead to better dehazing performance than its baseline AOD-Net.

%------------------------------------------------------------------------
\section{Related work}

In this section, we briefly summarize the sinlge-image dehazing methods that have been proposed in previous works and compare their advantages and deficiencies. Then, we proposed our perceptual guided end-to-end dehazing network that boosts the learning performance compared to the baseline AOD-Net.

The atmospheric scattering model has been widely used in previous haze removal work~\cite{mccartney1976optics,narasimhan2000chromatic,narasimhan2002vision}:

\begin{equation}
\label{haze model definition}
I(x) = J(x)t(x) + A(1-t(x)),
\end{equation}

\noindent where $x$ indexes pixels in the observed hazy image, $I(x)$ is observed hazing image, and $J(x)$ is the clean image to be recovered. The parameter $A$ denotes the global atmospheric light, and $t(x)$ is the transmission matrix defines as:

\begin{equation}
\label{trasmission matrix}
t(x) = e^{-\beta d(x)},
\end{equation}

\noindent where $\beta$ is the scattering coefficient of the atmosphere, and $d(x)$ represents the distance between the object and the camera.

The key to a successful haze removal algorithm is to recover the transmission matrix $t(x)$, on which the majority of the dehazing methods have focused through either physically grounded priors or data-driven approaches.

Conventional single image dehazing methods commonly exploit natural image priors and perform statical analysis. For example, \cite{he2011single,tang2014investigating} demonstrate that the dark channel prior (DCP) is informative to calculate the transmission matrix. \cite{zhu2015fast} proposed a color attenuation prior and created a linear model for scene depth of the hazy image to allow for an efficient supervised parameter learning method, and \cite{berman2016non} proposed a non-local prior based on the observation that pixels in a given cluster are often non-local and each color cluster in the clear image became a haze-line in RGB space.

More recently, CNNs have been applied to the haze removal application after demonstrated successes in many other computer vision tasks. \cite{ren2016single} exploits a multi-scale CNN~(MSCNN) that predicts a coarse-scale holistic transmission map of the entire image and refines it locally. \cite{cai2016dehazenet} proposed the \textit{DehazeNet}, a trainable transmission matrix estimator, and recovers the clean image combined with estimated global atmosphere light. Both these methods learn the transmission matrix from the CNN first and recover the haze-free image with separately calculated atmospheric light. Moreover, \cite{li2017reside} proposed a complete end-to-end dehazing network names AOD-Net which takes the hazy image as input and directly generates clean image output. 

In this project, we adopt the transformed atmospheric scattering model and the convolutional network architecture proposed in \cite{li2017aod} and aim at improving its performance by utilizing perceptually motivated loss functions.

%---------------------------------------------------
\section{Proposed work}

In this section, the proposed \textit{PAD-Net} is explained. We first introduce the transformed atmospheric scattering model and the dehazing network architecture design based on it, which we adopt the work in \cite{li2017aod} to facilitate an end-to-end single image dehazing. Then, we discuss the perceptually-motivated loss functions that will be explored in our project. 

\subsection{End-to-end Dehazing Network Design}
Based on the atmospheric scattering model (\ref{haze model definition}), the clean image generated by our network can be formulated as:
\begin{equation}
\label{transformed formula}
\begin{split}
J(x) = K(x)I(x) - K(x) + b, where \\
K(x) = \frac{\frac{1}{t(x)}(I(x)-A)+(A-b)}{I(x)-1},
\end{split}
\end{equation}
where $b$ is the constant bias whose default value is set to 1. Here, the core idea is to unify the two parameters in (\ref{haze model definition}) $t(x)$ and $A$ into one formula, i.e. $K(x)$, and directly minimize the reconstruction errors in the image pixel domain. Since $K(x)$ is dependent on the input $I(x)$, we in fact build an \textit{input-adaptive} deep model, and train the model by minimizing the reconstruction errors between its output $J(x)$ and ground truth clean image.

\begin{figure*}
\centering
\includegraphics[width=7.2in,height=1.7in]{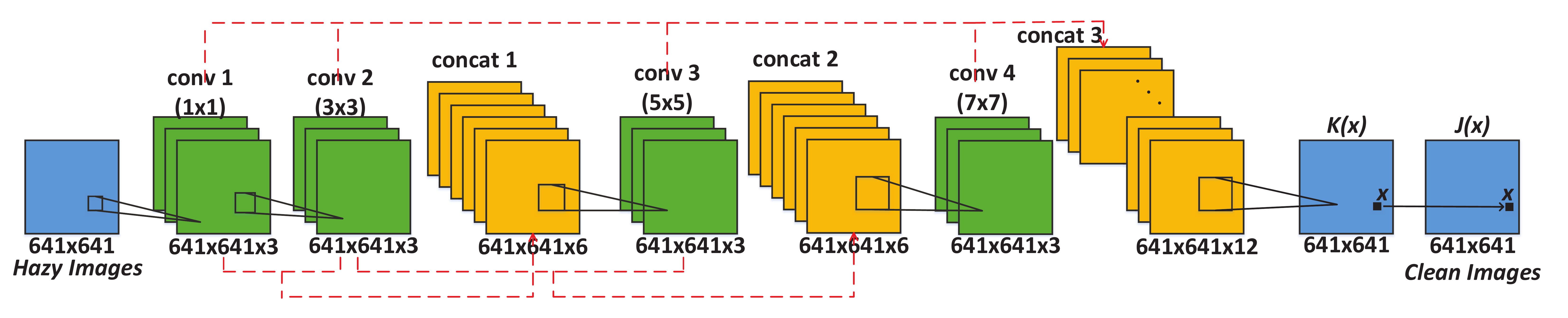}
\caption{The network diagram of PAD-Net}
\label{network_architecture}
\end{figure*}

Therefore, the proposed deep neural network is composed of two major parts: a K-estimation module to estimate $K(x)$ in (\ref{transformed formula}) with five convolutional layers, and a clean image generation modules that follows to produce the recovery clean image via element-wise calculation. The entire network diagram of the PAD-Net is visualized in Fig.~\ref{network_architecture}. 

As depicted in Fig.~\ref{network_architecture}, the five covolutional layers are implemented with different filter sizes to capture multi-scale features of the input hazy image and are concatenated with intermediates layers in order to compensate the information loss during convolutions, which is inspired by~\cite{cai2016dehazenet,ren2016single}. Output images (i.e., $J(x)$ in (\ref{transformed formula})) from the network is then compared with the ground truth clean image at the loss layer to compute the error function for back propagation. This end-to-end dehazing network can be easily embedded with other deep models as a stage in high-level computer vision tasks such as object detection, classification as well.

%Can add few sentences to justify, from analysis of the hazy model complexity. 
One thing to mention is that, the PAD-Net, inherited from the AOD-Net~\cite{li2017aod}, is a light-weighted network which has only three convolutional filters. In fact, if we analyze the atmospheric scattering model~((\ref{haze model definition}) and (\ref{trasmission matrix})), we can find that there are only three unknown parameters, $\beta$, $A$ and $d(x)$, in the model. And in our adopted benchmark, RESIDE dataset~\cite{li2017reside}, the $\beta$ and $A$ are pair-wised selected constants, and the depth map $d(x)$ are either calculated from a depth dataset such as NYU2~\cite{silberman2012indoor} or estimated with convolutional neural network~\cite{liu2016learning}. Therefore, the complexity of the hazy model is relatively low. Given this observation, in our work, we keep the setting of filter numbers in the AOD-Net to facilitate a quick training while obtaining good learning results.

\subsection{Perceptual Loss Functions}
\label{loss functions}
At the loss layer, different error functions will be investigated to optimize the image dehazing results and the results will be compared. In the following sections we introduce the error metrics that will be examined in our project. We show their key features and how to compute their derivative for backpropagation steps. These loss functions will be implemented in the loss layer individually or jointly, specified in Section \ref{evaluation}.

\begin{enumerate}[leftmargin=*]

%L2 norm
\item The $\ell_2$ error 

The $\ell_2$ norm of the error is generally chosen as the loss function for image dehazing \cite{li2017aod} given its simplicity and convexity. The $\ell_2$ norm penalizes large error, but is more tolerant to small error, regardless of the underlying structure in the image. As a result, it sometimes produces visible splotchy artifacts on the restored images. The HVS, on the other hand, is more sensitive to luminance and color variations in texture-less regions~\cite{winkler2004visibility}. The loss function for a patch $P$ can be written as:

\begin{equation}
\label{l2 error function}
\mathcal{L}^{\ell_2}(P) = \frac{1}{N}\sum_{p \in P}^{} (x(p)-y(p))^2,
\end{equation}

where $N$ is the number of pixels in the patch, $p$ is the index of the pixel, and $x(p)$ and $y(p)$ are the pixel values of the generated image and the ground truth image respectively. Since $\partial \mathcal{L}^{\ell_2}(P)/\partial q = 0,\forall q\neq p.$, for each pixel $p$ in the patch, the derivate can be denoted as:

\begin{equation}
\partial \mathcal{L}^{\ell_2}(P)/\partial x(p) = x(p) - y(p).
\end{equation}

Note that, even though $\mathcal{L}^{\ell_2}(P)$ is a function of the patch as a whole, the derivatives are back-propagated for each pixel in the patch.

%L1 norm
\item The $\ell_1$ error

The $\ell_1$ error is studied as an attempt to reduce the artifacts introduced by the $\ell_2$ and bring different convergence properties. Unlike the $\ell_2$ norm, the $\ell_1$ norm does not over-penalize large errors. The error function of $\ell_1$ is:
\begin{equation}
\label{l1 error function}
\mathcal{L}^{\ell_1}(P) = \frac{1}{N}\sum_{p \in P}^{} |x(p)-y(p)|.
\end{equation}

The derivatives of the $\ell_1$ is also simple. Similar to $\ell_2$ norm, the derivatives of a certain pixel in a patch only depends on the difference between its own value and the ground truth value at the same location and do not rely on other pixels in the same patch.

\begin{equation}
\label{l1 error function}
\partial \mathcal{L}^{\ell_1}(P)/\partial x(p) = sign(x(p) - y(p)).
\end{equation}

The derivative of $\mathcal{L}^{\ell_1}(P)$ is not defined at $0$. However, if the error is $0$, we do not need to update the weight. So here we use the convention that $sign(0)=0$.

\item SSIM
%give a better introduction on SSIM and MS-SSIM

Considering that image dehazing is a real-world application that reproduces visually clear and pleasing images, a perceptually motivated metric such as SSIM is worth studying. SSIM is a perception-based model that considers image degradation as perceived change in structural information, while also incorporating important perceptual phenomena, including both luminance masking and contrast masking terms. Inheriting the definition of $x(p)$ and $y(p)$ in (\ref{l2 error function}), and let $\mu_x$, ${\sigma_x}^2$, and $\sigma_{xy}$ be the mean of $x$, the variance of $x$, and the covariance of $x$ and $y$, approximately, $\mu_x$ and ${\sigma_x}$ can be viewed as estimates of the luminance and contrast of $x$, and $\mu_{xy}$ measures the structural similarity of $x$ and $y$ in terms of the tendency that they vary together. Then, the SSIM for pixel $p$ is defined as:

\begin{equation}
\label{SSIM}
\begin{split}
SSIM(P) & = \frac{2\mu_x\mu_y + C_1}{\mu_x^2+\mu_y^2+C_1}\cdot \frac{2\sigma_{xy}+C_2}{\sigma_x^2 + \sigma_y^2+C_2} \\
& =l(p)\cdot cs(p).
\end{split}
\end{equation}

\noindent where the means and standard deviations are computed with a Gaussian filer $G_{\sigma_G}$ with standard deviation $\sigma_G$. $l(p)$ and $cs(p)$ measure the comparisons of the luminance, and a combined contrast with structure similarity between $x$ and $y$ at the pixel $p$, respectively. The loss function for SSIM can be then defined as:
\begin{equation}
\label{SSIM error function}
\mathcal{L}^{SSIM}(P) = \frac{1}{N}\sum_{p \in P}^{} 1 - SSIM(p).
\end{equation}

Note that (\ref{SSIM}) indicates that the computation of SSIM($p$) requires looking at a neighboring of pixel $p$ since it involves the mean and standard deviations of the Gaussian filter $G_{\sigma_G}$ on the pixel. This means that $\mathcal{L}^{SSIM}(P)$, as well as its derivatives, cannot be calculated in some boundary region of P. 

However, the convolutional nature of the network studies in this work allows us to write the loss as:
\begin{equation}
\label{conv SSIM}
\mathcal{L}^{SSIM}(P) = 1 - SSIM(\widetilde p).
\end{equation}

\noindent where $\widetilde p$ is the center pixel of patch $P$. Again, this is because, even though the network learns the weights maximizing SSIM for the central pixel, the learned kernels are then applied to all the pixels int eh image. Note that the error can still be back-propagated to all the pixels within the support of $G_{\sigma_G}$ as they contribute to the computation of (\ref{conv SSIM}). More Formally, the derivatives of $\mathcal{L}^{SSIM}(P)$ at pixel $p$ can be computed as:

\begin{equation}
\begin{split}
\frac{\partial \mathcal{L}^{SSIM}(P)}{\partial x(q)} & =
-\frac{\partial}{\partial x(q)} SSIM(\widetilde p)\\
& = -\bigg( \frac{\partial l(\widetilde(p)}{\partial x(q)}\cdot cs(\widetilde p + 
l(\widetilde p)\cdot \frac{\partial cs(\widetilde p)}{\partial x(q)} \bigg),
\end{split}
\end{equation}

where $l(\widetilde p)$ and $cs(\widetilde p)$ are the first and second term of SSIM (i.e. (\ref{SSIM})) and their derivatives are:

\begin{equation}
\frac{\partial l(\widetilde(p)}{\partial x(q)} = 2\cdot G_{\sigma_G}(q - \widetilde p )\cdot \bigg( \frac{\mu_y - \mu_x\cdot l(\widetilde p)}{\mu_x^2+\mu_y^2+C_1}\bigg),
\end{equation}

\noindent and
\begin{equation}
\begin{split}
\frac{\partial cs(\widetilde(p)}{\partial x(q)} &= \frac{2}{\mu_x^2+\mu_y^2+C_2}\cdot G_{\sigma_G}(q-\widetilde p)\cdot [(y(q) - \mu_y) \\
& - cs(\widetilde p) \cdot (x(1) - \mu_x)],
\end{split}
\end{equation}

\noindent where $G_{\sigma_G}(q-\widetilde p)$ is the Gaussian coefficient associated with pixel $q$. 

\item MS-SSIM

The choice of $\sigma_G$ would impact training performance of SSIM. Specifically, the network with smaller $\sigma_G$ loses the ability to preserve the local structure and reintroduce splotchy artifacts in flat regions, while the network with large $\sigma_G$ tends to keep the noises in the proximity of edges. Our work, we adopt the idea of multi-scale SSIM from \cite{zhao2017loss} in which $M$ different values of $\sigma_G$ will be used rather than directly fine tune its value to optimize the performance of SSIM. Given a dyadic pyramid of $M$ levels, MS-SSIM is defined as:
\begin{equation}
\label{MS-SSIM}
\mathcal{L}^{MS-SSIM}(P) = l_M^\alpha(p) \cdot \prod_{j=1}^{M} cs_j^{\beta_j} (P).
\end{equation}

\noindent where $l_M$ and $cs_j$ are the terms design in (\ref{SSIM}) at scale $M$ and $j$, respectively. For convenience, we set $\alpha = \beta_j = 1$ for $j={1,..., M}$. Similarly to (\ref{conv SSIM}), we can approximate the loss for patch $P$ with the loss computed at its center pixel $\widetilde p$:

\begin{equation}
\label{conv MS-SSIM}
\mathcal{L}^{MS-SSIM}(P) = 1 - MS-SSIM(\widetilde p).
\end{equation}

The, the derivatives of the MS-SSIM loss function can be written as:

\begin{equation}
\begin{split}
& \frac{\partial \mathcal{L}^{MS-SSIM}(P)}{\partial x(q)} \\
& = \bigg ( \frac{\partial l_M (\widetilde p)}{\partial x(q)} + l_M(\widetilde P) \cdot \sum_{i=0}^{M} \frac{1}{cs_i(\widetilde p)} \frac{\partial cs_i (\widetilde p)}{\partial x(q)} \bigg) \cdot \prod_{j=1}^{M} cs_j(\widetilde p),
\end{split}
\end{equation}

To speed up the training, in stead of computing $M$ levels of pyramid $P$, we adopt the approach proposed in \cite{zhao2017loss} and ues $M$ different values for $\sigma_G$, each one being half of the previous, on the full-resolution patch. In the loss function defined in this work, ${\sigma_G}^i = {0.5, 1, 2, 4, 8}$. 

\end{enumerate}

%\section{Evaluation}
\section{Experimental Setup}
\label{evaluation}
\subsection{Systems}
%In our experiment, we are going to combine the AOD-Net with different loss functions. Specifically, we would like to examine the following systems,
In our experiment, the following error functions are applied to  the loss layer of the proposed \textit{PAD-Net}.
\begin{itemize}[noitemsep,topsep=2pt,parsep=2pt,partopsep=2pt]
\item \textbf{Baseline}: using $\ell_2$ loss alone
\item \textbf{L1}: using $\ell_1$ loss alone
\item \textbf{SSIM}: using SSIM loss alone
\item \textbf{MS-SSIM}: using MS-SSIM loss alone
\item \textbf{MS-SSIM+L2}: using a weighted sum of MS-SSIM and $\ell_2$ as the loss function:

\begin{equation}
\label{MSSSIM-L2}
\mathcal{L}^{MSSSIM-L2} = \alpha \cdot \mathcal{L}^{MSSSIM} + (1-\alpha)\cdot G_{\sigma_{G}^M} \cdot \mathcal{L}^{\ell_2},
\end{equation}

\noindent A point-wise multiplication between $G_{\sigma_{G}^M}$ and $\mathcal{L}^{\ell_2}$ is added for the $\ell_2$ loss function term because MS-SSIM propagates the error at pixel $q$ based on its contribution to MS-SSIM of the central pixel $\widetilde q$, as determined by the Gaussian weights.

\item \textbf{MS-SSIM+L1}: using a weighted sum of MS-SSIM and $\ell_1$ as the loss function:
\begin{equation}
\label{MSSSIM-L1}
\mathcal{L}^{MSSSIM-L1} = \alpha \cdot \mathcal{L}^{MSSSIM} + (1-\alpha)\cdot G_{\sigma_{G}^M} \cdot \mathcal{L}^{\ell_1},
\end{equation}

\noindent Similarly, the $\ell_1$ loss is also weighted by the Gaussian filter $G_{\sigma_{G}^M}$.

\end{itemize}
\subsection{Data and evaluation metrics}
The benchmark dataset for this project is the RESIDE dataset \cite{li2017reside}, which has a large number of pictures for training the neural network and evaluating the dehazing performance. The ITS (indoor images) and OTS (outdoor images) sets from RESIDE provide a total number of over 400,000 images for training. Due to the limits of time and computational resources, we randomly sampled 10,000 images from ITS and OTS as the training data for this work. Among the 10,000 images, 2,790 are from IST and the rest 7,210 are from OTS, and they roughly takes the same percentage of number of images in the corresponding dataset, respectively. We also randomly sampled another 1,000 non-overlapping set of images as the validation data. All proposed systems were evaluated on the held-out SOTS subset, which contains 1,000 synthetic haze images (500 indoor and 500 outdoor images).

We used PSNR and SSIM as objective measurements of dehazing performance. Due to the limited scope of this project, we will not be able to run subjective evaluations on the dehazing results.

\subsection{Implementation details}
\label{implementation}
We built our neural network and loss functions using PyCaffe \cite{jia2014caffe} because it has been proven to be flexible enough for research purposes while being able to support fast prototyping. The neural network architecture of AOD-Net was defined by following \cite{li2017aod}, and we referred to an open-source pre-trained model \cite{boyiliee} published by the original authors to make sure that our implementation was as close to the original AOD-Net as possible. Unless otherwise noted, the base learning rate and mini-batch size of the systems were set to 0.01 and 8, respectively. The networks were initialized using Gaussian random variables. We used a momentum of 0.9 and a weight decay of 0.0001, following \cite{li2017aod}. We also clipped the L2 norm of the gradient to be within [-0.1, 0.1] to stabilize the network training process, as suggested by \cite{pascanu2013difficulty}. All systems were trained on a Nvidia GTX 1070 GPU for around 14 epochs, which empirically ensures convergence.

\section{Results}
\label{results}
\subsection{Compare different loss functions}
\label{init-exp}

\begin{table}[t]
\begin{center}
\begin{tabular}{|l|c|c|c|}
\hline
\multirow{2}{*}{Systems}&\multicolumn{3}{c|}{PSNR}\\ \cline{2-4} 
&Indoor&Outdoor&All\\ \hline
AOD-Net&\textbf{21.01}&24.08&22.55\\ \hline
L2&20.73&25.58&23.15\\ \hline
L1&20.27&25.83&23.05\\ \hline
SSIM&19.64&26.65&23.15\\ \hline
MS-SSIM&19.54&\textbf{26.87}&23.20\\ \hline
MS-SSIM+L1&20.16&26.20&23.18\\ \hline
MS-SSIM+L2&20.45&26.38&\textbf{23.41}\\ \hline
\end{tabular}
\end{center}
\caption{PSNR results without fine-tuning}
\label{psnr-res-overall}
\end{table}

\begin{table}[t]
\begin{center}
\begin{tabular}{|l|c|c|c|}
\hline
\multirow{2}{*}{Systems}&\multicolumn{3}{c|}{SSIM}\\ \cline{2-4} 
&Indoor&Outdoor&All\\ \hline
AOD-Net&\textbf{0.8372}&0.8726&0.8549\\ \hline
L2&0.8235&0.9090&0.8663\\ \hline
L1&0.8045&0.9111&0.8578\\ \hline
SSIM&0.7940&0.8999&0.8469\\ \hline
MS-SSIM&0.8038&0.8989&0.8513\\ \hline
MS-SSIM+L1&0.8138&\textbf{0.9184}&0.8661\\ \hline
MS-SSIM+L2&0.8285&0.9177&\textbf{0.8731}\\ \hline
\end{tabular}
\end{center}
\caption{SSIM results without fine-tuning}
\label{ssim-res-overall}
\end{table}

In the first set of experiments, we would like to investigate that which loss function provides the best objective dehazing performance. For \textbf{SSIM}, the standard deviation of the Gaussian filter was set to $\sigma_G=5$. $C_1$ and $C_2$ in (\ref{SSIM}) were 0.01 and 0.03, respectively. For \textbf{MS-SSIM}, the Gaussian filters were constructed by setting $\sigma_G^i=\{0.5, 1, 2, 4, 8\}$. The \textbf{MS-SSIM+L1} loss function used $\alpha=0.025$, and \textbf{MS-SSIM+L2} used $\alpha=0.1$, following \cite{zhao2017loss}. The results are summarized in Table \ref{psnr-res-overall} and Table \ref{ssim-res-overall}. The AOD-Net's results were copied from the RESIDE dataset paper \cite{li2017reside}, which was trained using the whole OTS and ITS sets. To provide a fair comparison, we also used the L2 norm loss function to train on our training set. Overall, the system \textbf{MS-SSIM+L2} achieved the best performance on both PSNR and SSIM, outperformed AOD-Net by 3.8\% and 2.1\%, respectively. Although AOD-Net performed well on indoor images, we found that when the number of training samples was the same, \textbf{MS-SSIM+L2} achieved a similar PSNR (20.45) compared with \textbf{L2} (20.73) and a higher SSIM (0.8285 v.s. 0.8235) in the indoor case. In terms of performance on outdoor images, the proposed loss functions generally worked better than the AOD-Net.

\subsection{Fine-tuning \textit{MS-SSIM+L2}}
\begin{table}
\begin{center}
\begin{tabular}{|c|c|c|c|}
\hline
\multirow{2}{*}{$\alpha$} & \multicolumn{3}{c|}{PSNR}\\ \cline{2-4} 
&Indoor&Outdoor&All\\ \hline
0.1&\textbf{20.68}&26.18&23.43\\ \hline
0.3&20.47&26.49&23.48\\ \hline
0.5&20.46&26.39&23.43\\ \hline
0.7&20.32&\textbf{26.67}&\textbf{23.50}\\ \hline
0.9&20.50&26.34&23.42\\ \hline
\end{tabular}
\end{center}
\caption{PSNR results with fine-tuning on MS-SSIM+L2}
\label{psnr-res-finetune}
\end{table}

\begin{table}
\begin{center}
\begin{tabular}{|c|c|c|c|}
\hline
\multirow{2}{*}{$\alpha$} & \multicolumn{3}{c|}{SSIM}\\ \cline{2-4} 
&Indoor&Outdoor&All\\ \hline
0.1&\textbf{0.8229}&\textbf{0.9266}&\textbf{0.8747}\\ \hline
0.3&0.8197&0.9248&0.8722\\ \hline
0.5&0.8116&0.9226&0.8671\\ \hline
0.7&0.8140&0.9211&0.8676\\ \hline
0.9&0.8165&0.9204&0.8685\\ \hline
\end{tabular}
\end{center}
\caption{SSIM results with fine-tuning on MS-SSIM+L2}
\label{ssim-res-finetune}
\end{table}

\begin{figure}[t!]
\centering
\includegraphics[width=3.25in]{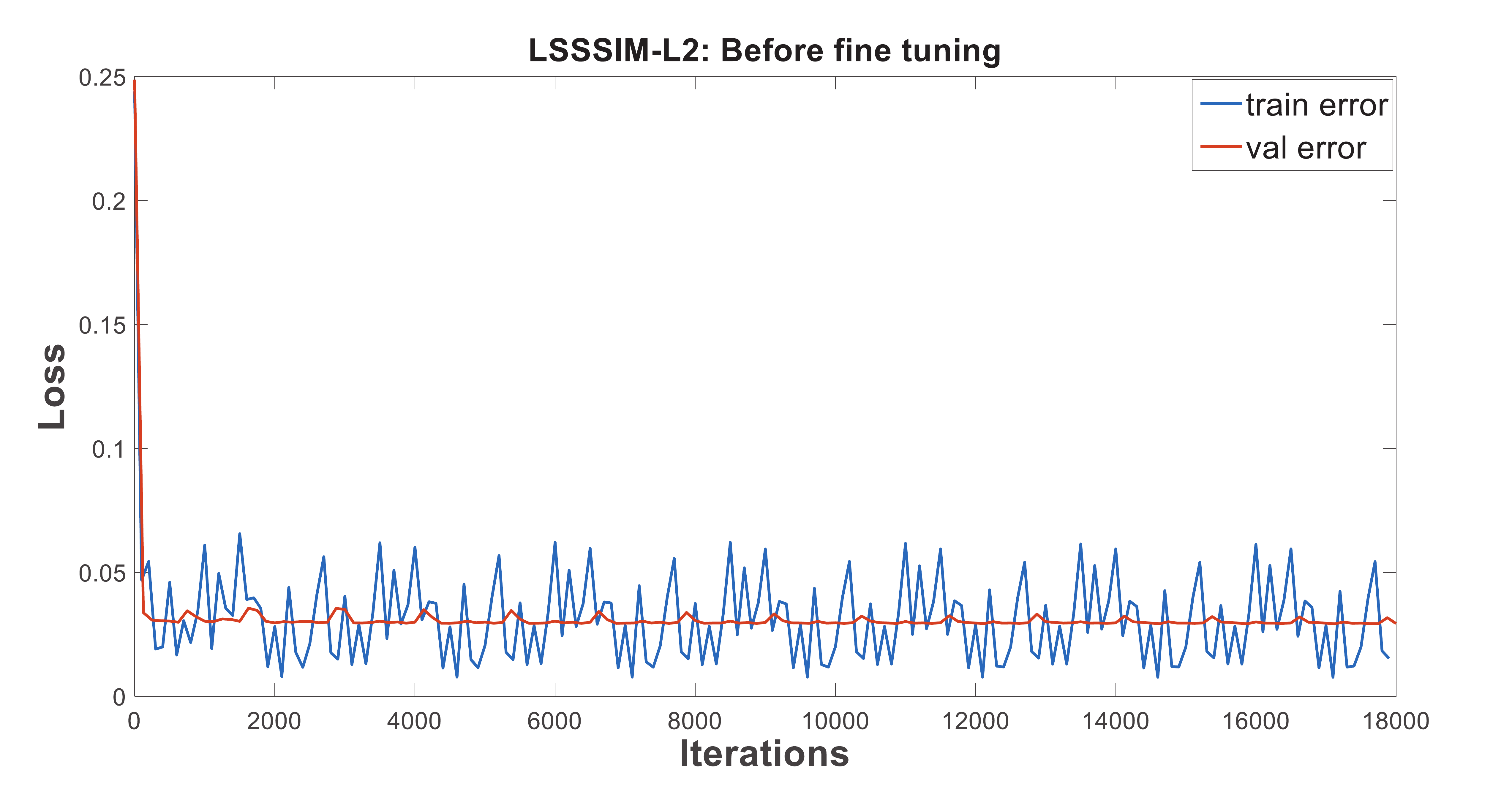}
\caption{A sample learning curve of MS-SSIM+L2 before fine-tuning, the training loss curve was sampled every 10 iterations.}
\label{learning_curve}
\end{figure}

\begin{figure}[t!]
\centering
\includegraphics[width=3.25in]{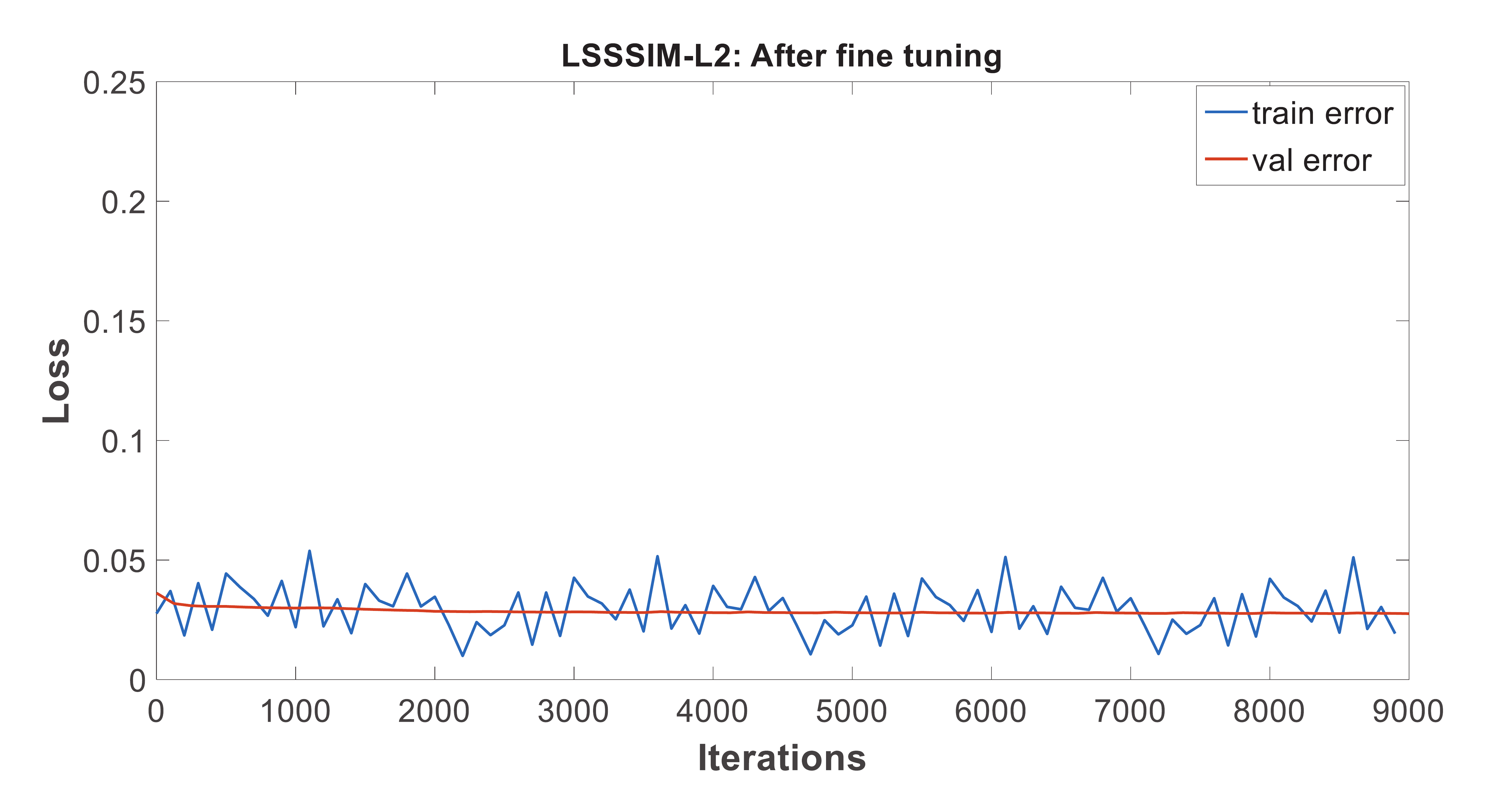}
\caption{A sample learning curve of MS-SSIM+L2 after fine-tuning with $\alpha=0.1$, the training loss curve was sampled every 10 iterations.}
\label{learning_curve_fine_tune}
\end{figure}

Based on the results above, we performed further fine-tuning on system \textbf{MS-SSIM+L2}, expecting to see better performance. We first analyzed the learning curve of the MISSIM-L2 system before fine-tuning, as illustrated in Fig.~\ref{learning_curve}. It can be seen from the plot that the learning curve converged quickly and the learning errors fluctuate from iterations to iterations, which indicated we may need a smaller learning rate and a larger mini-batch size. Given that, we used weights from a pre-trained AOD-Net model \cite{boyiliee} to initialize our network. During training, we applied a smaller learning rate (0.002) and a larger mini-batch size (16). We also tested on different $\alpha$ values to adjust the contribution of MS-SSIM and L2 to the fused loss function. All the other hyper-parameter were kept unchanged as in Section \ref{implementation}. The results are summarized in Tables \ref{psnr-res-finetune} and \ref{ssim-res-finetune}. For PSNR, setting $\alpha$ to 0.7 yielded the best performance (23.50). For SSIM, the highest score (0.8747) was achieved when setting $\alpha$ to 0.1. We also plot the learning curve after fine tuning in Fig.~\ref{learning_curve_fine_tune} and it demonstrates that after fine tuning, the error starts at a small value at the beginning of the training due to the pre-train process. And the learning curve is now smoother and flatter as a result of the better choice of the learning rate and mini-batch size.

\section{Discussion}

From the results in Section \ref{results} we can see that the proposed loss functions work well on outdoor images, but not as good on indoor testing samples. Part of the reason was that our training set contains limited number of outdoor samples, and the reported AOD-Net used around 40 times more indoor samples than our implementation for training. As we noted in Section \ref{init-exp}, when we used the same training set to train on the L2 norm and the proposed loss functions, we achieved similar (PSNR) or better (SSIM) performance on indoor testing samples, not to mention that we have significant higher outdoor dehazing performances.

Surprisingly, we found that by directly optimizing the SSIM loss function, we did not obtain the optimal SSIM performance on the test set. After taking a closer look at our implementation, we found that we used a different standard deviation (1.5) for the Gaussian filter in the evaluation SSIM function. When we set the standard deviation as 5 and re-ran the SSIM evaluation, the SSIM score became 0.8597. Though this result is not directly comparable with the other systems, it is larger than when using 1.5 as the standard deviation. We did not use standard deviation 1.5 for training the dehazing model because that will be "cheating" -- using a parameter from the evaluation system. Still, it would be interesting to see if setting standard deviation to 1.5 in training could produce optimal SSIM performance on the testing data.

The learning curve of the training process converged very fast, even after fine tuning, as depicted in Fig. \ref{learning_curve_fine_tune}. We suspected that this was due to the fact that the parameter set used for generating those synthetic images is small, therefore, the neural network might not need too many training iterations to figure out the value of those parameters.

The training costs, in terms of time, for some of the proposed cost functions (SSIM, MS-SSIM) were significant higher than using the $\ell_2$ norm. For one thing, computing the gradient of those cost functions were of higher time complexity, since the derivate of the $\ell_2$ error only involves the computation on the pixel itself while SSIM and MS-SSIM needs to do Gaussian filtering among the patch, as explained in Section \ref{loss functions} . For another, the $\ell_2$ norm implementation was carefully optimized to take full advantages of GPU computing, whereas our implementation was experimental and built on PyCaffe's python layers, which was recognized for having flexibility for fast prototyping but slow in training. We believe that further implementation optimization can bridge the gap of training cost between the $\ell_2$ norm loss and the other loss functions.

\begin{figure}[t!]
\centering
\includegraphics[width=3.25in]{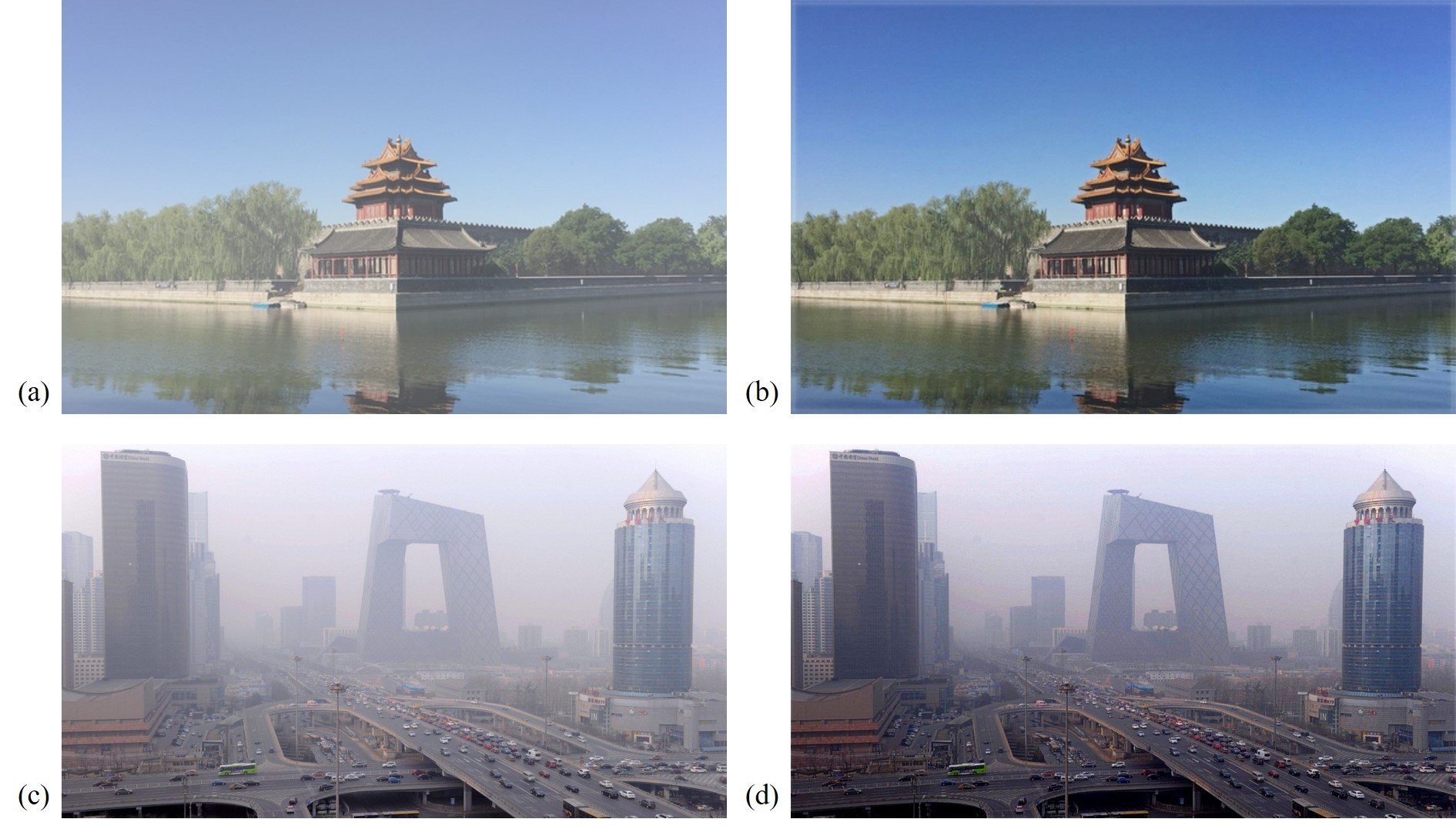}
\caption{Dehazing examples. (a) A synthetic haze image. (b) Dehazed version of (a). (c) A real haze image. (d) Dehazed version of (c)}
\label{dehaze_examples}
\end{figure}

Fig.~\ref{dehaze_examples} shows some dehaze example using the model trained on the MS-SSIM+L2 loss function ($\alpha=0.1$). The haze images are from the HTS set in RESIDE. We can see that even though the dehazing neural network was trained only on synthetic data, it can successfully remove haze from both synthetic and real haze images.

\section{Conclusion}

In this project, we propose to use perception-motivated loss functions to train an end-to-end dehazing neural network. Compared with a baseline system that has the same neural network architecture but uses the conventional $\ell_2$ norm (MSE) loss function, we obtained significantly better objective dehazing performance on the SOTS set on the RESIDE dataset. The best PSNR we obtained was 23.50 (4.2\% relative improvement), and the best SSIM we obtained was 0.8747 (2.3\% relative improvement). For future work, first, we would like to train the proposed system using the full OTS and ITS sets. Second, we will investigate system MS-SSIM+L1 more closely -- we did not have time to fine-tune the parameters of this system, but the initial results were still promising. Third, we are planning to conduct perceptual studies to ask human raters to judge the quality of the dehazed images.

In terms of logistics, the workload devision was fair among the two teammates, the details are summarized in Table \ref{workload}. The project code is open-source, accessible on GitHub at \url{https://github.com/guanlongzhao/single-image-dehazing}.

\begin{table}
\begin{center}
\begin{tabular}{|l|c|c|}
\hline
Action items&Liu&Zhao\\ \hline
Project idea&60\%&40\%\\ \hline
System implementation&40\%&60\%\\ \hline
Experiment and data analysis&40\%&60\%\\ \hline
Report and presentation&60\%&40\%\\ \hline
\end{tabular}
\end{center}
\caption{Project management}
\label{workload}
\end{table}

{\small
\bibliographystyle{ieeetr}
\bibliography{csce633_report}
}

\end{document}